\documentclass[10pt,twocolumn,letterpaper]{article}

\usepackage{cvpr}
\usepackage{times}
\usepackage{epsfig}
\usepackage{graphicx}
\usepackage{amsmath}
\usepackage{amssymb}
\usepackage{url}
\usepackage{subfigure}
\usepackage{amsmath}
\usepackage{amsthm}
\usepackage{amssymb}
\usepackage{algorithmic}
\usepackage{algorithm}
\usepackage{epstopdf}
\usepackage{cite}
\usepackage{multirow}
\usepackage{setspace}

\usepackage[pagebackref=true,breaklinks=true,letterpaper=true,colorlinks,bookmarks=false]{hyperref}

\cvprfinalcopy 

\def\cvprPaperID{3430} 

\ifcvprfinal\fi
\begin{document}

\title{NestedNet: Learning Nested Sparse Structures in Deep Neural Networks}

\author{Eunwoo Kim ~~\qquad \qquad  Chanho Ahn \qquad \qquad~~ Songhwai Oh\\
Department of ECE and ASRI, Seoul National University, South Korea \\
{\tt\small \{kewoo15,  mychahn, songhwai\}@snu.ac.kr}
}

\maketitle
\thispagestyle{empty}

\begin{abstract}

Recently, there have been increasing demands to construct compact deep
architectures to remove unnecessary redundancy and to improve the
inference speed.
While many recent works focus on reducing the redundancy by
eliminating unneeded weight parameters, it is not possible to apply a
single deep network for multiple devices with different
resources.
When a new device or circumstantial condition requires a new deep
architecture, it is necessary to construct and train a new network
from scratch.
In this work, we propose a novel deep learning framework, called a
nested sparse network, which exploits an $n$-in-1-type nested
structure in a neural network.
A nested sparse network consists of multiple levels of networks with
a different sparsity ratio associated with each level, and higher level
networks share parameters with lower level networks to enable stable
nested learning.
The proposed framework realizes a resource-aware versatile
architecture as the same network can meet diverse resource
requirements, i.e., anytime property.
Moreover, the proposed nested network can learn different forms of
knowledge in its internal networks at different levels, enabling
multiple tasks using a single network, such as coarse-to-fine
hierarchical classification.
In order to train the proposed nested network, we propose
efficient weight connection learning and channel and layer scheduling
strategies.
We evaluate our network in multiple tasks, including adaptive deep
compression, knowledge distillation, and learning class hierarchy, and
demonstrate that nested sparse networks perform competitively, but
more efficiently, compared to existing methods.

\end{abstract}

\section{Introduction}\label{sec:intro}

\begin{figure}[t]
    \centering
    {\includegraphics[width=0.43\textwidth]{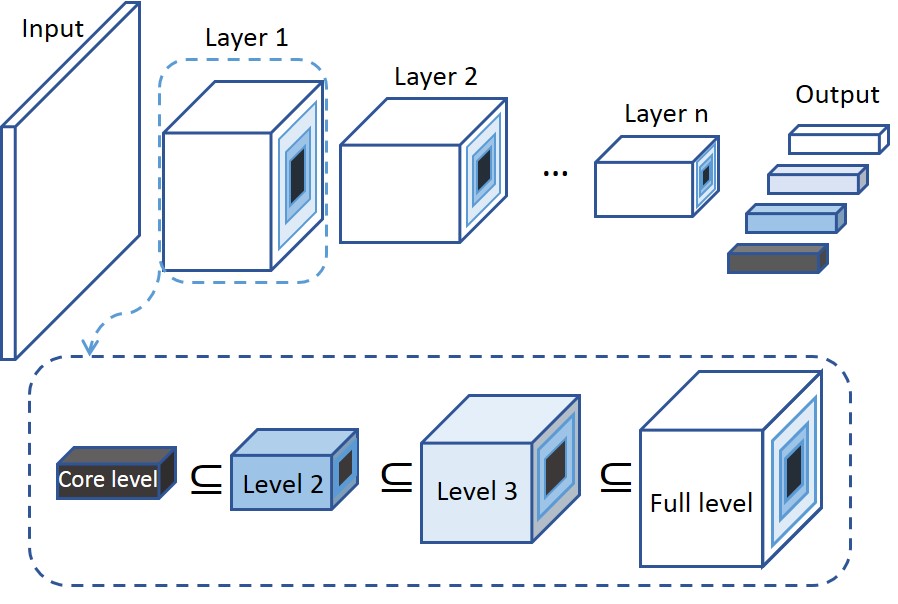}}
    \caption{
    Conceptual illustration of the nested sparse network with $n$ nested levels ($n$-in-$1$ network, $n$=$4$ here).
    A nested sparse network consists of internal networks from core level (with the sparsest parameters) to full level (with all parameters)
    and an internal network share its parameters with higher level networks, making a network-in-network structure.
    Since the nested sparse network produces multiple different outputs,
    it can be leveraged for multiple tasks as well as multiple devices with different resources.}
    \label{fig:nsn_basic}
\end{figure}

Deep neural networks have recently become a standard architecture
due to their significant performance improvement
over the traditional machine learning models in a number of fields,
such as
image recognition \cite{krizhevsky2012imagenet, he2016deep, huang2017densely},
object detection \cite{ren2015faster},
image generation \cite{goodfellow2014generative}, and
natural language processing \cite{bengio2003neural, sutskever2014sequence}.
The successful outcomes are derived from the availability of massive labeled data
and computational power to process such data.
What is more, many studies have been conducted toward very deep and dense models
\cite{simonyan2015very, he2016deep, zagoruyko2016wide, huang2017densely}
to achieve further performance gain.
Despite of the success, the remarkable progress is accomplished at the expense of
intensive computational and memory requirements,
which can limit deep networks for a practical use,
especially on mobile devices with low computing capability.
In particular, if the size of a network architecture is designed to be colossal,
it may be problematic for the network to achieve mission-critical
tasks on a commercial device which requires a real-time operation.

Fortunately, it is well-known that there exists much redundancy in
most of deep architectures, i.e., a few number of network
parameters represent the whole deep network in substance
\cite{lecun1990optimal}.
This motivates many researchers to exploit the redundancy from multiple points of view.
The concept of sparse representation is to elucidate the redundancy
by representing a network with a small number of representative parameters.
Most of sparse deep networks prune connections with insignificant
contributions \cite{han2015learning, han2016deep, zhou2016less, alvarez2016learning, yang2017designing, liu2017learning},
prune the number of channels \cite{li2017pruning, he2017channel, liu2017learning},
or prune the number of layers \cite{wen2016learning} by sparse regularization.
Another regularization strategy to eliminate the network redundancy is low-rank approximation
which approximates weight tensors by minimizing the reconstruction error between the original network
and the reduced network
\cite{jaderberg2014speeding, tai2015convolutional, zhang2016accelerating, kim2016compression, zhou2016less}.
Weight tensors can be approximated by decomposing into tensors of pre-specified sizes
\cite{jaderberg2014speeding, tai2015convolutional, zhang2016accelerating, kim2016compression}
or by solving a nuclear-norm regularized optimization problem \cite{zhou2016less}.

Obviously, developing a compact deep architecture is beneficial
to satisfy the specification of a device with low capacity.
However, it is difficult for a learned compact network to be adjusted
for different hardware specifications (e.g., different sparsity levels),
since a deep neural network normally learns parameters for a given task.
When a new model or a device with a different computing budget is required,
we usually define a new network again manually by trial and error.
Likewise, if a different form of knowledge is required in a trained network,
it is hard to keep the learned knowledge while training using the same
network again or using a new network \cite{hinton2015distilling}.
In general, to perform multiple tasks, we need multiple networks at
the cost of considerable computation and memory footprint.

In this work, we aim to exploit a nested structure in a deep neural
architecture which realizes an $n$-in-1 versatile network to conduct
multiple tasks within a single neural network (see Figure~\ref{fig:nsn_basic}).
In a nested structure, network parameters are assigned to multiple
sets of nested levels, such that a low level set is a subset of
parameters to a higher level set.
Different sets can capture different forms of knowledge according to
the type (or amount) of information,
making it possible to perform multiple tasks using a single network.

To this end, we propose a nested sparse network, termed \textit{NestedNet},
which consists of multiple levels of networks with different sparsity
ratios (nested levels), where an internal network with lower nested
level (higher sparsity) shares its parameters with other internal
networks with higher nested levels in a network-in-network fashion.
Thus, a lower level internal network can learn common knowledge while
a higher level internal network can learn task-specific knowledge.
It is well-known that
early layers of deep neural networks share general knowledge
and later layers learn task-specific knowledge.
A nested network learns more systematic hierarchical representation
and has an effect of grouping analogous filters
for each nested level as shown in Section \ref{exp:deep_comp}.
NestedNet also enjoys another useful property,
called anytime property \cite{zilberstein1996using, larsson2016fractalnet},
and it can produce early (coarse) prediction with a low level network and
more accurate (fine) answer with a higher level network.
Furthermore, unlike existing networks, the nested sparse network can learn different forms
of knowledge in its internal networks with different levels.
Hence, the same network can be applied to multiple tasks satisfying
different resource requirements,
which can reduce the efforts to train separate existing networks.
In addition, consensus of different knowledge in a nested network can
further improve the performance of the overall network.
In order to exploit the nested structure,
we present several pruning strategies
which can be used to learn parameters from scratch using off-the-shelf deep learning libraries.
We also provide applications, in which the nested structure can be applied,
such as adaptive deep compression, knowledge distillation, and
hierarchical classification.
Experimental results demonstrate that NestedNet
performs competitively compared to popular baseline and other
existing sparse networks.
In particular, our results in each application (and each data) are produced
from a single nested network, making NestedNet highly efficient
compared with currently available approaches.

In summary, the main contributions of this work are:
\begin{itemize}
\item
We present an efficient connection pruning method, which learns
sparse connections from scratch. We also provide channel and layer
pruning by scheduling to exploit the nested structure to avoid the
need to train multiple different networks.
\item
We propose an $n$-in-1 nested sparse network to realize the nested structure in a
deep network.
The nested structure enables not only resource-aware anytime prediction
but knowledge-aware adaptive learning for various tasks which are not
compatible with existing deep architectures.
Besides, consensus of multiple knowledge can improve the prediction of NestedNet.
\item
The proposed nested networks are performed on various applications in
order to demonstrate its efficiency and versatility at
comparable performance.
\end{itemize}

\section{Related Work}

A na\"{\i}ve approach to compress a deep neural network is to prune network connections
by sparse approximation.
Han et al. \cite{han2015learning} proposed an iterative prune and retrain approach using the $l_1$- or $l_2$-norm regularization.
Zhou et al. \cite{zhou2016less} proposed a forward-backward splitting method to solve the $l_1$-norm regularized optimization problem.
Note that the weight pruning methods with non-structured sparsity can be difficult
to achieve valid speed-up using standard machines due to their irregular memory access \cite{wen2016learning}.
Channel pruning approaches were proposed
by structured sparsity regularization \cite{wen2016learning}
and channel selection methods \cite{he2017channel, liu2017learning}.
Since they reduce the actual number of parameters, they have benefits of computational and memory resources
compared to the weight connection pruning methods.
Layer pruning \cite{wen2016learning} is another viable approach for compression
when the parameters associated with a layer has little contributions
in a deep neural network using short-cut connection \cite{he2016deep}.
There is another line of compressing deep networks by low-rank approximation,
where weight tensors are approximated by low-rank tensor decomposition
\cite{jaderberg2014speeding, tai2015convolutional, zhang2016accelerating, kim2016compression}
or by solving a nuclear-norm regularized optimization problem \cite{zhou2016less}.
It can save memory storage and enable valid speed-up when learning and inferencing the network.
The low-rank approximation approaches, however, normally require a pre-trained model
when optimizing parameters to reduce the reconstruction error with the original learned parameters.

It is important to note here that
the learned networks using the above compression approaches
are difficult to be utilized for different tasks,
such as different compression ratios,
since the learned parameters are trained for a single task (or a
given compression ratio).
If a new compression ratio is required,
one can train a new network with manual model parameter tuning from scratch or
further tune the trained network to suit the new demand\footnote{Additional tuning
on a trained network with a new requirement
can be accompanied by forgetting the learned knowledge \cite{li2017learning}.},
and this procedure will be conducted continually whenever the form of the model is changed,
requiring additional resources and efforts.
This difficulty can be fully addressed using the proposed nested sparse network.
It can embody multiple internal networks within a network and
perform different tasks at the cost of learning a single network.
Furthermore, since the nested sparse network is constructed from scratch,
the effort to learn a baseline network is not needed.

There have been studies to build a compact network
from a learned large network,
called knowledge distillation,
while maintaining the knowledge of the large network \cite{hinton2015distilling, adriana2015fitnets}.
It shares intention with the deep compression approaches
but utilizes the teacher-student paradigm to ease the training of networks \cite{adriana2015fitnets}.
Since it constructs a separate student network from a learned teacher network,
its efficiency is also limited similar to deep compression models.

The proposed nested structure is also related to tree-structured deep architectures.
Hierarchical structures in a deep neural network have been recently
exploited for improved learning
\cite{yan2015hd, larsson2016fractalnet, kim2017splitnet}.
Yan et al. \cite{yan2015hd} proposed a hierarchical architecture that
outputs coarse-to-fine predictions using different internal networks.
Kim et al. \cite{kim2017splitnet} proposed a structured deep network that
can enable model parallelization and a more compact model compared
with previous hierarchical deep networks.
However, since their networks do not have a nested structure since
parameters in their networks form independent groups in the hierarchy,
they cannot have the benefit of nested learning for sharing
knowledge obtained from coarse- to fine-level sets of parameters.
This limitation is discussed more in Section \ref{exp:taxonomy}.

\section{Compressing a Neural Network}

Given a set of training examples $X = [x_1, x_2, ..., x_n]$ and
labels $Y = [y_1, y_2, ..., y_n]$,
where $n$ is the number of samples,
a neural network learns a set of parameters $\mathcal{W}$ by minimizing the following optimization problem
\begin{equation}\label{eq:dl_basic}
\min_{\mathcal{W}}~  \mathcal{L}\Big(Y, f(X, \mathcal{W})\Big) + \lambda \mathcal{R}(\mathcal{W}),
\end{equation}
where $\mathcal{L}$ is a loss function between the network output and the ground-truth label,
$\mathcal{R}$ is a regularizer which constrains weight parameters,
and $\lambda$ is a weighting factor balancing between loss and regularizer.
$f(X, \mathcal{W})$ outputs according to the purpose of a task, such as classification (binary number) and regression (real number),
through a chain of linear and nonlinear operations using the parameter $\mathcal{W}$.
A set of parameters is represented by $\mathcal{W} = \{W_l\}_{1\leq l \leq L}$,
where $L$ is the number of layers in a network,
and $W_l \in \mathbb{R}^{k_w \times k_h \times c_i \times c_o}$ for a convolutional weight
or $W_l \in \mathbb{R}^{c_i \times c_o}$ for a fully-connected weight
in popular deep learning architectures such as AlexNet \cite{krizhevsky2012imagenet},
VGG networks \cite{simonyan2015very}, and residual networks \cite{he2016deep}.
Here, $k_w$ and $k_h$ are the width and height of a convolutional kernel and
$c_i$ and $c_o$ are the number of input and output channels (or activations\footnote{Activations denote neurons in fully-connected layers.}), respectively.

In order to exploit a sparse structure in a neural network,
many studies usually try to enforce constraints on $\mathcal{W}$,
such as sparsity using the $l_1$ \cite{han2015learning, zhou2016less}
or $l_2$ weight decay \cite{han2015learning}
and low-rank-ness using the nuclear-norm \cite{zhou2016less}
or tensor factorization \cite{jaderberg2014speeding, yu2017compressing}.
However, many previous studies utilize a pre-trained network
and then prune connections in the network
to develop a parsimonious network,
which usually requires significant additional computation.

\section{Nested Sparse Networks}\label{sec:nsn}

\subsection{Sparsity learning by pruning}\label{sec:nsn-1}

We investigate three pruning approaches for sparse deep learning:
(entry-wise) weight connection pruning, channel pruning, and layer pruning,
which are used for nested sparse networks described in Section \ref{sec:nsn-2}.

To achieve weight connection pruning,
pruning strategies were proposed to reduce learned parameters using a pre-defined threshold \cite{han2015learning} and
using a subgradient method \cite{zhou2016less}.
However, they require additional pruning steps to sparsify a learned
dense network.
As an alternative, we propose an efficient sparse connection learning approach
which learns from scratch without additional pruning steps using the
standard optimization tool.
The problem formulation can be constructed as follows:
\begin{equation}\label{eq:our_sdl}
\begin{split}
\min_{\mathcal{W}}~ \mathcal{L}\Big(Y, f(&X, \mathcal{P}_{\Omega_{\mathcal{M}}}(\mathcal{W}))\Big)
+ \lambda \mathcal{R}\Big(\mathcal{P}_{\Omega_{\mathcal{M}}}(\mathcal{W})\Big)\\
&\mbox{s.t.} ~~ \mathcal{M} \triangleq \sigma(\alpha(\mathcal{W}) - \tau),
\end{split}
\end{equation}
where $\mathcal{P}(\cdot)$ is the projection operator
and $\Omega_{\mathcal{M}}$ denotes the support set of $\mathcal{M} = \{M_l\}_{1\leq l \leq L}$.
$\alpha(\cdot)$ is the element-wise absolute operator and
$\sigma(\cdot)$ is an activation function to encode binary output (such as the unit-step function) and
$\tau$ is a pre-defined threshold value for pruning.
Since the unit-step function makes learning the problem (\ref{eq:our_sdl}) by standard back-propagation difficult due to its discontinuity,
we present a simple approximated unit-step function:
\begin{equation}\label{eq:approx_unit_step}
\sigma(x) = tanh(\gamma \cdot x_+),
\end{equation}
where $x_+ = max(x, 0)$, 
$tanh(\cdot)$ is the hyperbolic tangent function,
and $\gamma$ is a large value to mimic the slope of the unit-step function.\footnote{We set $\gamma = 10^5$ and it
is not sensitive to initial values of parameters
when applying the popular initialization method \cite{glorot2010understanding}
from our empirical experiences.}
Note that $\mathcal{M}$ acts as an implicit mask of $\mathcal{W}$ to reveal sparse weight tensor.
Once an element of $\mathcal{M}$ becomes 0,
its corresponding weight is no longer updated in the optimization procedure,
making no more contribution to the network.
By solving (\ref{eq:our_sdl}), we construct a sparse deep network
based on off-the-shelf deep learning libraries without additional efforts.

To achieve channel or layer pruning\footnote{Layer pruning can be applicable to
the structure in which weights of the same size are repeated, such as residual networks \cite{he2016deep}.},
we consider the following weight (or channel) scheduling problem:
\begin{equation}\label{eq:ch_prune}
\min_{\mathcal{W}}~ \mathcal{L}\Big(Y, f(X, \mathcal{P}_{\Omega_{\mathcal{M}'}}(\mathcal{W}))\Big)
+ \lambda  \mathcal{R}\Big(\mathcal{P}_{\Omega_{\mathcal{M}'}}(\mathcal{W})\Big),
\end{equation}
where $\mathcal{M}' = \{M_l'\}_{1 \leq l \leq L} \subseteq \mathcal{M}$ consists of binary weight tensors
whose numbers of input and output channels (or activations for fully-connected layers)
are reduced to smaller numbers than the numbers of channels (activations) in the baseline architecture
to fulfill the demanded sparsity.
In other words, we model a network with a single number of scheduled channels using $\mathcal{M}'$
and then optimize $\mathcal{W}$ in the network from scratch.
Achieving multiple sparse networks by scheduling multiple numbers of channels is described in the following section.
Similar to the channel pruning, implementing layer pruning is straight-forward by reducing the number of layers
in repeated blocks \cite{he2016deep}.
In addition, pruning approaches can be combined for various nested
structures as described in the next section.

\subsection{Nested sparse networks}\label{sec:nsn-2}

The goal of a nested sparse network is to represent
an $n$-in-1 nested structure of parameters in a deep neural network
to allow $n$ nested internal networks as shown in Figure \ref{fig:nsn_basic}.
In a nested structure,
an internal network with lower (resp., higher) nested level gives higher (resp., lower) sparsity on parameters,
where higher sparsity means a smaller number of non-zero entries.
In addition, the internal network of the core level (resp., the lowest level) defines
the most compact sub network among the internal networks
and the internal network of the full level (resp., the highest level) defines the fully dense network.
Between them, there can be other internal networks with intermediate sparsity ratios.
Importantly, an internal network of a lower nested level shares its parameters
with other internal networks of higher nested levels.

\begin{figure}[t]
    \centering
    {\includegraphics[width=0.45\textwidth]{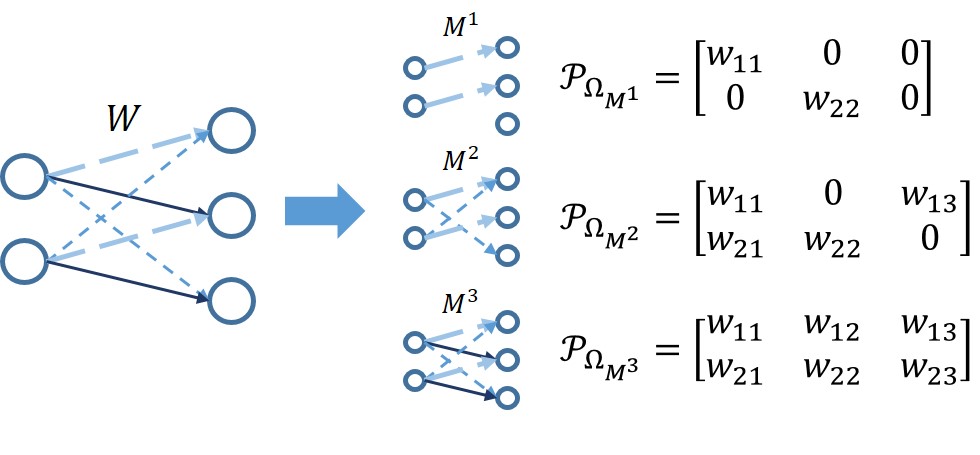}}
    \caption{
    A graphical representation of nested parameters $W$ by masks with three levels between fully-connected layers.
    }
    \label{fig:nsn_fc}
\end{figure}

Given a set of masks $\mathcal{M}$,
a nested sparse network,
where network parameters are assigned to multiple sets of nested levels,
can be learned by optimizing the following problem:
\begin{equation}\label{eq:nsn}
\setlength{\jot}{0.5pt}
\begin{split}
\min_{\mathcal{W}} \frac{1}{l_n}  \bigg( \sum_{j=1}^{l_n} &\mathcal{L}\Big(Y, f(X, \mathcal{P}_{\Omega_{\mathcal{M}^j}}(\mathcal{W}))\Big) \bigg)
+ \lambda  \mathcal{R} \Big(\mathcal{P}_{\Omega_{\mathcal{M}^{l_n}}}(\mathcal{W})\Big) \\
\mbox{s.t.}&~~  \mathcal{M}^j \subseteq \mathcal{M}^k, ~ j \leq k, ~ \forall j,k \in [1, ..., l_n],
\end{split}
\end{equation}
where $l_n$ is the number of nested levels.
Since a set of masks $\mathcal{M}^j = \{M_l^j\}_{1\leq l \leq L}$ represents the set of $j$-th nested level weights by its binary values,
$\mathcal{P}_{\Omega_{\mathcal{M}^j}}(\mathcal{W}) \subseteq \mathcal{P}_{\Omega_{\mathcal{M}^k}}(\mathcal{W}), j \leq k, \forall j,k \in [1, ..., l_n]$.
A simple graphical illustration of nested parameters between fully-connected layers is shown in Figure \ref{fig:nsn_fc}.
By optimizing (\ref{eq:nsn}), we can build a nested sparse network with nested levels by $\mathcal{M}$.

In order to find a set of masks $\mathcal{M}$,
we apply three pruning approaches described in Section \ref{sec:nsn-1}.
First, for realizing a nested structure in entry-wise weight connections,
masks are estimated by solving the weight connection pruning problem (\ref{eq:our_sdl})
with $l_n$ different thresholds iteratively.
Specifically, once the mask $\mathcal{M}^k$ consisting of the $k$-th nested level weights
is obtained in a network\footnote{Since we use the approximation in (\ref{eq:approx_unit_step}),
an actual binary mask is obtained by additional thresholding after the mask is estimated.},
we further train the network from the masked weight $\mathcal{P}_{\Omega_{\mathcal{M}^k}}(\mathcal{W})$
using a higher value of threshold to get another mask giving a higher sparsity,
and this procedure is conducted iteratively until reaching the sparsest mask of the core level.
This strategy is helpful to find sparse dominant weights \cite{han2015learning},
and our network trained using this strategy performs better
than a network whose sparse mask is obtained randomly
and other sparse networks in Section \ref{exp:deep_comp}.

\begin{figure}[t]
    \centering
    {\includegraphics[width=0.49\textwidth]{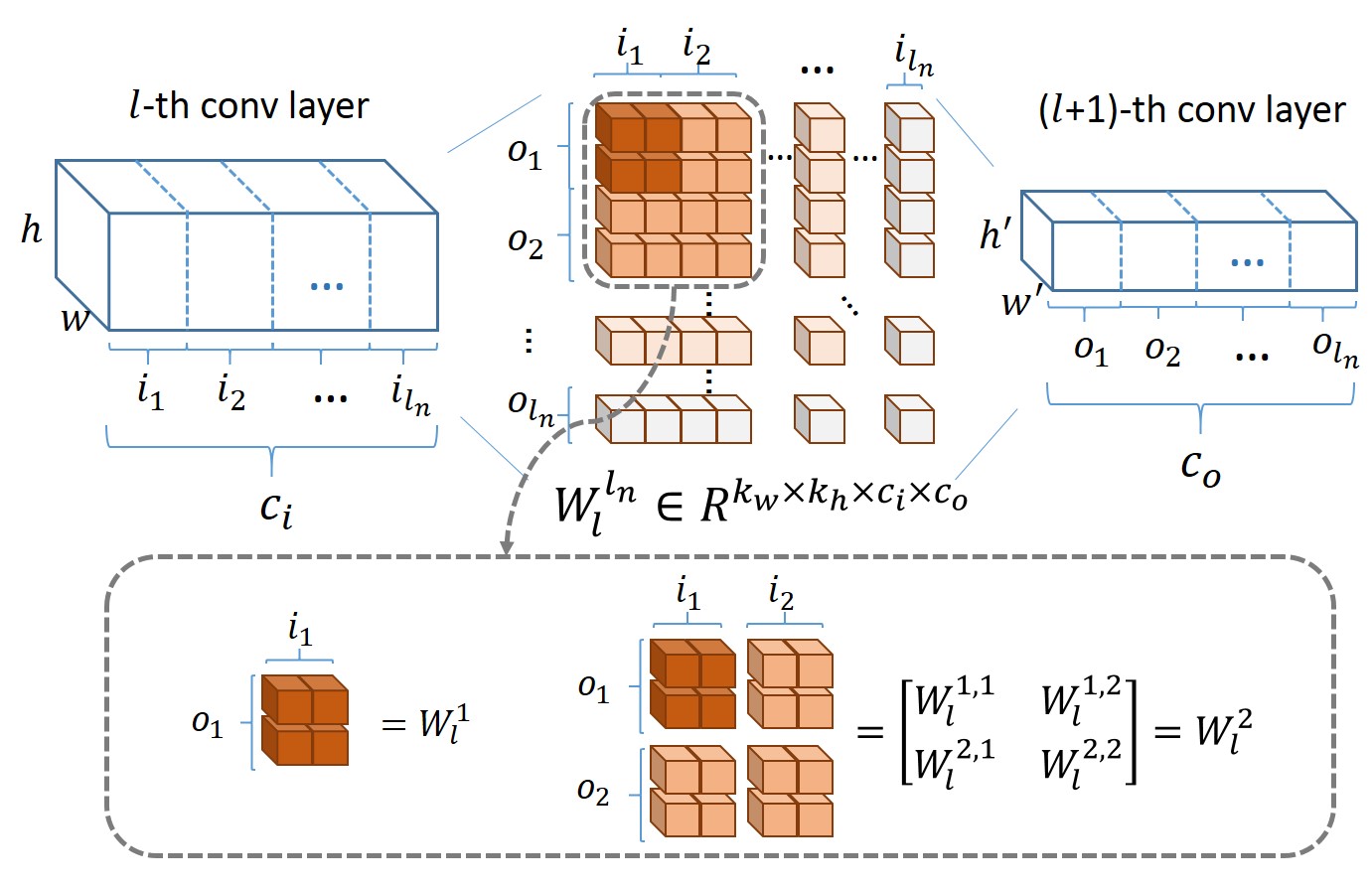}}
    \caption{
    A graphical representation of channel scheduling using the nested parameters
    between $l$-th and ($l+1$)-th convolutional layers in a nested sparse network
    where both layers are scheduled.
    $i_k$ and $o_k$ denote additional numbers of input and output channels
    in the $k$-th level, respectively.}
    \label{fig:nsn_sch}
\end{figure}

For a nested sparse structure in convolutional channels or layers,
we schedule a set of masks $\mathcal{M}$ according to the type of pruning.
In the channel-wise scheduling,
the number of channels in convolutional layers and
the dimensions in fully-connected layers are scheduled
to pre-specified numbers for all scheduled layers.
The scheduled weights are learned by solving (\ref{eq:nsn})
without performing the mask estimation phase in (\ref{eq:our_sdl}).
Mathematically, we represent weights from the first (core) level weight $W_l^1 \in \mathbb{R}^{k_w \times k_h \times i_1 \times o_1}$ to
the full level weight $W_l = W_l^{l_n} \in \mathbb{R}^{k_w \times k_h \times c_i \times c_o}$,
where $c_i = \sum_m i_m$ and $c_o = \sum_m o_m$,
between $l$-th and ($l+1$)-th convolutional layers as
\begin{equation}\label{eq:nsn_sch}
\setlength{\jot}{1.5pt}
\begin{aligned}
&W_l^1 = W_l^{1,1}, \\
&W_l^2 =
\begin{bmatrix}
W_l^{1,1} & W_l^{1,2} \\ W_l^{2,1} & W_l^{2,2} \\
\end{bmatrix},\\
& \quad \quad  \vdots \\
& W_l = W_l^{l_n} =
\begin{bmatrix}
W_l^{1,1} & W_l^{1,2} & ... & W_l^{1,l_n} \\
W_l^{2,1} & W_l^{2,2} & ... & W_l^{2,l_n} \\
\vdots & \vdots & \vdots & \vdots \\
W_l^{l_n,1} & W_l^{l_n,2} & ... & W_l^{l_n,l_n}
\end{bmatrix}.
\end{aligned}
\end{equation}
Figure \ref{fig:nsn_sch} illustrates the nested sparse network with channel scheduling,
where different color represents weights in different nested level except its shared sub-level weights.
For the first input layer, observation data is not scheduled in this work
(i.e., $c_i$ is not divided).
Unlike the nested sparse network with the weight pruning method,
which holds a whole-size network structure for any nested levels,
channel scheduling only keeps and learns the parameters corresponding to the number of scheduled channels
associated with a nested level,
making valid speed-up especially for an inference phase.

Likewise, we can schedule the number of layers and its corresponding weights
in a repeated network block and learn parameters by solving (\ref{eq:nsn}).
Note that for a residual network which consists of $6 n_b + 2$ layers \cite{he2016deep},
where $n_b$ is the number of residual blocks,
if we schedule $n_b = \{2,3,5\}$,
our single nested residual network with $l_n=3$ consists of three residual networks of size 14, 20, and 32 in the end.
Among them, the full level network with $n_b=5$ has the same number of parameters to the conventional residual network of size 32
without introducing further parameters.

\subsection{Applications}\label{sec:nsn-3}


\noindent \textbf{Adaptive deep compression.}
Since a nested sparse network is constructed under the weight connection learning,
it can apply to deep compression \cite{han2016deep}.
Furthermore, the nested sparse network realizes adaptive deep compression
because of its anytime property \cite{zilberstein1996using} which makes it possible to
provide various sparse networks and
infer adaptively using a learned internal network
with a sparsity level suitable for the required specification.
For this problem, we apply weight pruning and channel scheduling
presented in Section \ref{sec:nsn-2}.
\vspace{2mm}

\begin{figure}[t]
    \centering
    {\includegraphics[width=0.45\textwidth]{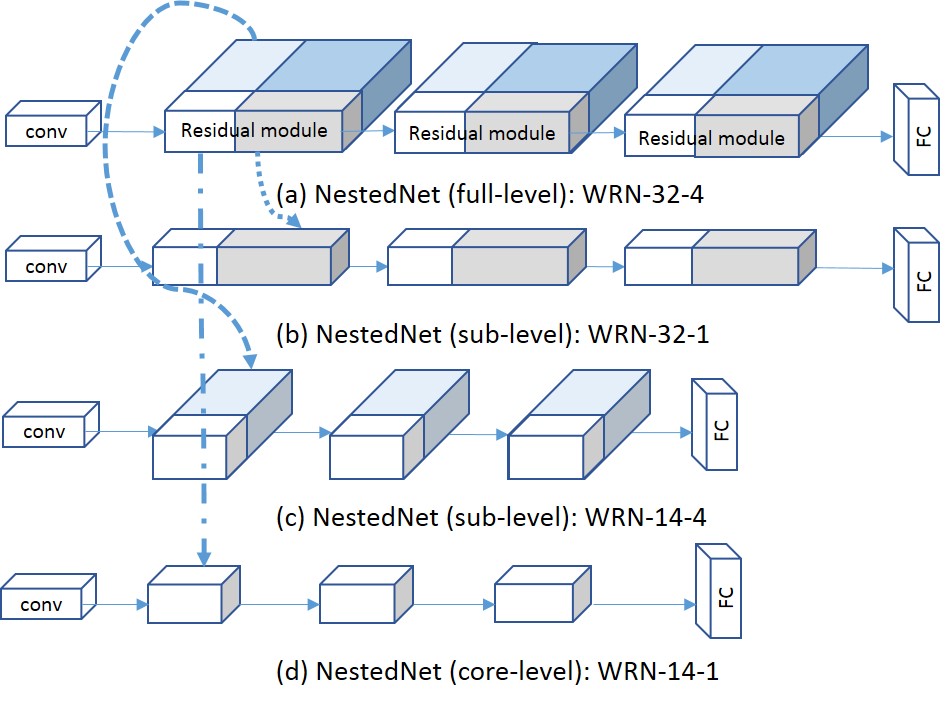}}
    \caption{
    NestedNet with four nested levels
    using channel and layer scheduling for knowledge distillation.}
    \label{fig:kd_cifar100}
\end{figure}

\noindent \textbf{Knowledge distillation.}
Knowledge distillation is used to represent knowledge compactly in a network \cite{hinton2015distilling}.
Here, we apply channel and layer scheduling approaches to make small-size sub-networks as shown in Figure \ref{fig:kd_cifar100}.
We train all internal networks, one full-level and $l_n-1$ sub-level networks,
simultaneously from scratch without pre-training the full-level network.
Note that the nested structure in sub-level networks may not be necessarily
coincided with the combination of channel and layer scheduling
(e.g., subset constraint is not satisfied for Figure~\ref{fig:kd_cifar100} (b) and (c))
according to a design choice. \vspace{2mm}

\noindent \textbf{Hierarchical classification.}
In a hierarchical classification problem \cite{yan2015hd},
a hierarchy can be modeled as an internal network with a nested level.
For example, we model a nested network with two nested levels for the CIFAR-100 dataset \cite{krizhevsky2009learning}
as it has 20 super classes, where each class has 5 subclasses (a total 100 subclasses).
It enables nested learning to perform coarse to fine representation and inference.
We apply the channel pruning method since it can handle different output dimensionality.

\section{Experiments}\label{sec:exp}

We have evaluated NestedNet
based on popular deep architectures,
ResNet-$n$ \cite{he2016deep} and WRN-$n$-$k$ \cite{zagoruyko2016wide},
where $n$ is the number of layers and $k$ is the scale factor on the number of convolutional channels
for the above three applications.
Since it is difficult to compare fairly with other baseline and sparse networks
due to their non-nested structure,
we provide a one-to-one comparison between internal networks in NestedNet
and their corresponding independent baselines
or other published networks of the same network structure.
NestedNet was performed on three benchmark datasets:
CIFAR-10, CIFAR-100 \cite{krizhevsky2009learning},
and ImageNet \cite{krizhevsky2012imagenet}.
The test time is computed for a batch set of the same size to the training phase.
All NestedNet variants and other compared baselines were implemented
using the TensorFlow library \cite{abadi2016tensorflow}
and processed by an NVIDIA TITAN X graphics card.
Implementation details of our models are described in Appendix.

\subsection{Adaptive deep compression}\label{exp:deep_comp}

We applied the weight connection and channel pruning approaches described in Section \ref{sec:nsn-1}
based on ResNet-56 to compare with the state-of-the-art network (weight connection) pruning \cite{han2015learning}
and channel pruning approaches \cite{li2017pruning, he2017channel}.
We implemented the iterative network pruning method for \cite{han2015learning}
under our experimental environment giving the same baseline accuracy,
and results of channel pruning approaches \cite{li2017pruning, he2017channel}
under the same baseline network were refereed from \cite{he2017channel}.
To compare with the channel pruning approaches,
our nested network was constructed with two nested levels,
full-level (1$\times$ compression) and core-level (2$\times$ compression),
and to compare with the network pruning method,
we constructed another NestedNet with three internal networks
(1$\times$, 2$\times$, and 3$\times$ compressions)
by setting $\boldsymbol{\tau} = (\tau_{1\times}, \tau_{2\times}, \tau_{3\times}) = (0, 0.015, 0.025)$,
where the full-level networks give the same result to the baseline network, ResNet-56 \cite{he2016deep}.
In the experiment, we also provide results of the three-level nested network
(1$\times$, 2$\times$, and 3$\times$ compressions),
which is learned using random sparse masks,
in order to verify the effectiveness of the proposed weight connection learning method in Section \ref{sec:nsn-1}.

\begin{table}[t]
\caption{Deep compression results using ResNet-56 for the CIFAR-10 dataset.
($\cdot$) denotes the compression rate of parameters from the baseline network.
Baseline results are obtained by author's implementation without pruning (1$\times$).}
\centering
    \renewcommand\arraystretch{1.0}
    \small
    \begin{tabular}{ |c ||  c | c |} \hline
     Method                           & Accuracy  & Baseline    \\ \hline \hline
     Filter pruning  \cite{li2017pruning} (2$\times$)   & 91.5$\%$  & 92.8$\%$    \\
     Channel pruning \cite{he2017channel} (2$\times$)   & 91.8$\%$  & 92.8$\%$    \\
     NestedNet - channel pruning (2$\times$)    & 92.9$\%$  & 93.4$\%$    \\ \hline \hline
     Network pruning \cite{han2015learning} (2$\times$)           & 93.4$\%$   & 93.4$\%$     \\
     NestedNet - random mask (2$\times$)        & 91.2$\%$  & 93.4$\%$    \\
     NestedNet - weight pruning (2$\times$)     & 93.4$\%$  & 93.4$\%$    \\ \hline
     Network pruning \cite{han2015learning} (3$\times$)           & 92.6$\%$   & 93.4$\%$     \\
     NestedNet - random mask (3$\times$)        & 85.1$\%$  & 93.4$\%$    \\
     NestedNet - weight pruning (3$\times$)     & 92.8$\%$  & 93.4$\%$    \\
    \hline
    \end{tabular}
    \label{tab:deep_comp}
\end{table}

\begin{figure*}[t]
    \centering
    \subfigure[]
    {\includegraphics[width=0.42\textwidth]{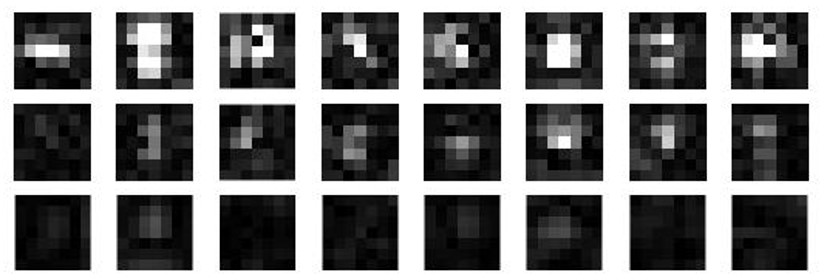}\quad \quad
    \label{fig:deep_compress_filters}}
    \subfigure[]
    {\includegraphics[width=0.38\textwidth]{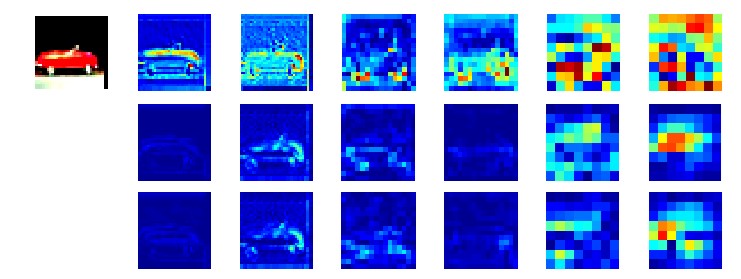}
    \label{fig:deep_compress_act}}
    \caption{
    Results from NestedNet on the CIFAR-10 dataset.
    (a) Learned filters from the first convolutional layer.
    (b) Activation feature maps from a train image (upper left).
    Each column represents a different layer (the layer number increases from left to right).
    Each row represents the images in each internal network,
    from core-level (top) to full-level (bottom).
    Best viewed in color.
    }
    \label{fig:nsn_compress}
\end{figure*}

Table \ref{tab:deep_comp} shows the classification accuracy of the compared networks for the CIFAR-10 dataset.
For channel pruning, NestedNet gives smaller performance loss from baseline
than recently proposed channel pruning approaches \cite{li2017pruning, he2017channel}
under the same reduced parameters (2$\times$),
even though the baseline performance is not the same
due to their different implementation strategies.
For weight connection pruning, ours performs better than network pruning \cite{han2015learning} on average.
They show no accuracy compromise under 2$\times$ compression,
but ours gives better accuracy than \cite{han2015learning}
under 3$\times$ compression.
Here, the weight connection pruning approaches outperform
channel pruning approaches including our channel scheduling based network under 2$\times$ compression,
since they prune unimportant connections in element-wise
while channel pruning approaches eliminate connections
in group-wise (thus dimensionality itself is reduced)
which can produce information loss.
Note that the random connection pruning gives the poor performance,
confirming the benefit of the proposed connection learning approach in learning the nested structure.

Figure \ref{fig:deep_compress_filters} represents learned filters
(brighter represents more intense)
of the nested network with the channel pruning approach using ResNet-56
with three levels (1$\times$, 2$\times$, and 4$\times$ compressions)
where the size of the first convolutional filters were set to $7 \times 7$ to see
observable large size filters under the same performance.
As shown in the figure, the connections in the core-level internal network are dominant
and upper-level filters,
which do not include their sub-level filters (when drawing the filters),
have lower importance than core-level filters which may learn side information of the dataset.
We also provide quantitative results for filters in three levels
using averaged normalized mutual information for all levels:
$0.38 \pm 0.09$ for within-level and $0.08 \pm 0.03$ for between-level,
which reveal that the nested network learns more diverse filters between nested levels than within levels
and it has an effect of grouping analogous filters for each level.
Figure \ref{fig:deep_compress_act} shows the activation maps (layer outputs)
of an image for different layers.
For more information, we provide additional activation maps for both train and test images
in Appendix.

\subsection{Knowledge distillation}\label{exp:kd}

To show the effectiveness of nested structures,
we evaluated NestedNets using channel and layer scheduling for knowledge distillation
where we learned all internal networks jointly
rather than learning a distilled network from a pre-trained model
in the literature \cite{hinton2015distilling}.
The proposed network was constructed under the WRN architecture \cite{zagoruyko2016wide}
(here WRN-32-4 was used).
We set the full-level network to WRN-32-4 and applied
(1) channel scheduling with $\frac{1}{4}$ scale factor (WRN-32-1 or ResNet-32),
(2) layer scheduling with $\frac{2}{5}$ scale factor (WRN-14-4),
and (3) combined scheduling for both channel and layer (WRN-14-1 or ResNet-14).
In the scenario, we did not apply the nested structure for the first convolutional layer and
the final output layer.
We applied the proposed network to CIFAR-10 and CIFAR-100.

\begin{figure}[t]
    \centering
    {\includegraphics[width=0.49\textwidth]{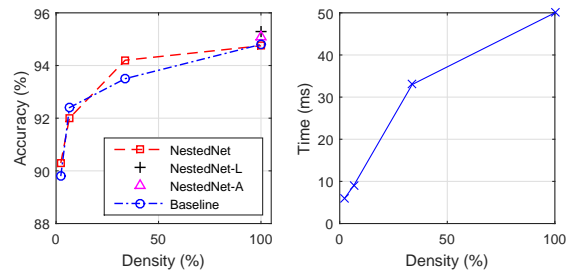}}
    \caption{
    Knowledge distillation results w.r.t accuracy (left) and time (right)
    using NestedNet for the CIFAR-10 dataset.}
    \label{fig:kd_cifar10_res}
\end{figure}

Figure \ref{fig:kd_cifar10_res} shows the comparison between NestedNet
with four internal networks, and their corresponding baseline networks learned independently
for the CIFAR-10 dataset.
We also provide test time of every internal network.\footnote{Since
the baseline networks require the same number of parameters and time as our networks,
we just present test time of our networks.}
As observed in the figure,
NestedNet performs competitively compared to its baseline networks for most of the density ratios.
Even though the total number of parameters to construct the nested sparse network
is smaller than that to learn its independent baseline networks,
the shared knowledge among the multiple internal networks
can compensate for the handicap and give the competitive performance to the baselines.
When it comes to test time, we achieve valid speed-up
for the internal networks with reduced parameters,
from about 1.5$\times$ (37$\%$ density) to 8.3$\times$ speed-up (2.3$\%$ density).
Table \ref{tab:knowledge_distill} shows the performance of NestedNet,
under the same baseline structure to the previous example, for the CIFAR-100 dataset.
$N_C$ and $N_P$ denote the number of classes and parameters, respectively.
In the problem, NestedNet is still comparable to its corresponding baseline networks on average,
which requires similar resource to the single baseline of full level.

\begin{table*}[t]
\caption{
Knowledge distillation results on the CIFAR-100 dataset.
${\star}$ indicates that the approximated total resource for the nested network.
}
\centering
    \renewcommand\arraystretch{1.0}
    \small
    \begin{tabular}{ |c || c |  c |  c | c | c | c | c || c |} \hline
     Network & Architecture   & $N_P$ & Density   & Memory & Accuracy & Baseline & Test time & Consensus ($N_C=100$) \\ \hline \hline
     \multirow{4}{*}{NestedNet} & WRN-32-4 & ~~7.4M$^{\star}$ & 100$\%$  & ~~28.2MB$^{\star}$   & 75.5$\%$ & 75.7$\%$ & 52ms (1$\times$) &   \\
      & WRN-14-4   & 2.7M & 37$\%$   & 10.3MB & 74.3$\%$ & 73.8$\%$ & 35ms (1.5$\times$) & NestedNet-A: 77.4$\%$ \\
      & WRN-32-1   & 0.47M & 6.4$\%$    & 1.8MB & 67.1$\%$ & 67.5$\%$ & 10ms (5.2$\times$) & NestedNet-L: 78.1$\%$ \\
      & WRN-14-1   & 0.18M & 2.4$\%$    & 0.7MB & 64.3$\%$ & 64.0$\%$ & 7ms  (7.4$\times$) & \\
    \hline
    \end{tabular}
    \label{tab:knowledge_distill}
\end{table*}

\subsection{Hierarchical classification}\label{exp:taxonomy}

We evaluated the nested sparse network for hierarchical classification.
We first constructed a two-level nested network for the CIFAR-100 dataset,
which consists of two-level hierarchy of classes,
and channel scheduling was applied to handle different dimensionality in the hierarchical structure of the dataset.
We compared with the state-of-the-art architecture,
SplitNet \cite{kim2017splitnet}, which can address class hierarchy.
Following the practice in \cite{kim2017splitnet},
NestedNet was constructed under WRN-14-8
and we adopted WRN-14-4 as a core internal network (4$\times$ compression).
Since the number of parameters in SplitNet is reduced to nearly 68$\%$ from the baseline,
we constructed another NestedNet based on the WRN-32-4 architecture
which has the almost same number of parameters as SplitNet.

Table \ref{tab:taxonomy} shows the performance comparison among the compared networks.
Overall, our two NestedNets based on different architectures
give the better performance than their baselines for all cases,
since ours can learn rich knowledge from
not only learning the specific classes but learning their abstract level (super-class) knowledge within the nested network,
compared to merely learning independent class hierarchy.
NestedNet also outperforms SplitNet for both architectures.
While SplitNet learns parameters which are divided into independent sets,
NestedNet learns shared knowledge for different tasks which can further improve the performance
by its combined knowledge obtained from multiple internal networks.
The experiment shows that the nested structure can realize encompassing multiple semantic knowledge in a single network
to accelerate learning.
Note that if the number of internal networks increases for more hierarchy,
the amount of resources saved increases.

\begin{table}[t]
\caption{
Hierarchical classification results on the CIFAR-100 dataset.
${\star}$ indicates that the approximated total resource for the nested network.}
\centering
    \renewcommand\arraystretch{1.0}
    \small
    \begin{tabular}{ |c || c | c | c | c | } \hline
     Network & Architecture & $N_C$ & $N_P$   & Accuracy    \\ \hline \hline
     \multirow{2}{*}{Baseline}  & WRN-14-4 & 20    & 2.7M   & 82.4$\%$  \\
                                & WRN-14-8 & 100   & 10.8M   & 75.8$\%$   \\ \hline
     \multirow{2}{*}{NestedNet} & WRN-14-4   & 20    & 2.7M     & 83.9$\%$  \\
                                & WRN-14-8   & 100   & ~~10.8M$^{\star}$  & 77.3$\%$   \\ \hline
     NestedNet-A &   WRN-14-8   & 100 & ~~10.9M$^{\star}$ & 78.3$\%$  \\ \hline
     NestedNet-L &   WRN-14-8   & 100 & ~~10.9M$^{\star}$ & 78.0$\%$  \\ \hline \hline
     SplitNet \cite{kim2017splitnet} & WRN-14-8 & 100   & 7.4M    &  74.9$\%$   \\ \hline \hline
     \multirow{2}{*}{Baseline}  & WRN-32-2 & 20    & 1.8M  & 82.1$\%$    \\
                                & WRN-32-4 & 100   & 7.4M   & 75.7$\%$    \\  \hline
     \multirow{2}{*}{NestedNet} & WRN-32-2 & 20    & 1.8M    & 83.7$\%$  \\
                                & WRN-32-4  & 100   & ~~7.4M$^{\star}$ & 76.6$\%$  \\ \hline
     NestedNet-A &   WRN-32-4   & 100 & ~~7.4M$^{\star}$ & 78.0$\%$ \\ \hline
     NestedNet-L &   WRN-32-4   & 100 & ~~7.4M$^{\star}$ & 77.7$\%$  \\ \hline

    \end{tabular}
    \label{tab:taxonomy}
\end{table}

We also provide experimental results
on the ImageNet (ILSVRC 2012) dataset \cite{imagenet_cvpr09}.
From the dataset, we collected a subset,
which consists of 100 diverse classes
including natural objects, plants, animals, and artifacts.
We constructed a three-level hierarchy
and the numbers of super and intermediate classes are 4 and 11, respectively
(a total 100 subclasses).
Taxonomy of the dataset is summarized in Appendix.
The number of train and test images are
128,768 and 5,000, respectively,
which were collected from the original ImageNet dataset \cite{imagenet_cvpr09}.
NestedNet was constructed based on the ResNet-18 architecture
following the instruction in \cite{he2016deep} for the ImageNet dataset,
where the numbers of channels in the core and intermediate level networks
were set to quarter and half of the number of all channels, respectively,
for every convolutional layer.
Table \ref{tab:imagenet} summarizes the hierarchical classification results
for the ImageNet dataset.
The table shows that NestedNet,
whose internal networks are learned simultaneously in a single network,
outperforms its corresponding baseline networks for all nested levels.

\subsection{Consensus of multiple knowledge}\label{exp:consensus}

One important benefit of NestedNet is to leverage multiple knowledge
of internal networks in a nested structure.
To utilize the benefit, we appended another layer at the end,
which we call a consensus layer,
to combine outputs from all nested levels for more accurate prediction by
1) averaging (\textit{NestedNet-A}) or
2) learning (\textit{NestedNet-L}).
For NestedNet-L,
we simply added a fully-connected layer
to the concatenated vector of all outputs in NestedNet,
where we additionally collected the fine class output in the core level network
for hierarchical classification.
See Appendix for more details.
While the overhead of combining outputs of different levels of NestedNet is negligible,
as shown in the results for knowledge distillation and hierarchical classification,
the two consensus approaches outperform the existing structures
including NestedNet of full-level under the similar number of parameters.
Notably, NestedNet of full-level in hierarchical classification gives better performance
than that in knowledge distillation under the same architecture, WRN-32-4,
since it has rich knowledge by incorporating coarse class information
in its architecture without introducing additional structures.

\begin{table}[t]
\caption{
Hierarchical classification results on ImageNet.}
\centering
    \renewcommand\arraystretch{1.0}
    \small
    \begin{tabular}{ | c || c | c | c |  } \hline
     Network &  $N_C$ & $N_P$  & Accuracy  \\ \hline \hline
     \multirow{3}{*}{Baseline}  & 4    & 0.7M  &  92.8$\%$  \\
                                & 11   & 2.8M  &  89.2$\%$  \\
                                & 100  & 11.1M &  79.8$\%$  \\ \hline
     \multirow{3}{*}{NestedNet} & 4    & 0.7M  &  94.0$\%$  \\
                                & 11   & 2.8M  &  90.2$\%$  \\
                                & 100  & 11.1M &  79.9$\%$  \\ \hline
     NestedNet-A  &  100  & 11.1M &  80.2$\%$   \\ \hline
     NestedNet-L  &  100  & 11.1M &  80.3$\%$  \\ \hline
    \end{tabular}
    \label{tab:imagenet}
\end{table}

\section{Conclusion}\label{sec:conclusion}

We have proposed a nested sparse network, named NestedNet,
to realize an $n$-in-1 nested structure in a neural network,
where several networks with different sparsity ratios are contained
in a single network and learned simultaneously.
To exploit such structure,
novel weight pruning and scheduling strategies have been presented.
NestedNet is an efficient architecture to incorporate multiple knowledge
or additional information within a neural network,
while existing networks are difficult to embody such structure.
NestedNets have been extensively tested on various applications
and demonstrated that it performs competitively,
but more efficiently,
compared to existing deep architectures.
\vspace{3mm}

{\small
\noindent \textbf{Acknowledgements: }
This research was supported in part
by Basic Science Research Program through the National Research Foundation of Korea (NRF)
funded by the Ministry of Science, ICT $\&$ Future Planning (NRF-2017R1A2B2006136),
by the Next-Generation Information Computing Development Program
through the National Research Foundation
of Korea (NRF) funded by the Ministry of Science and ICT (2017M3C4A7065926), and
by the Brain Korea 21
Plus Project in 2018.
}

{\small
\bibliographystyle{ieee}
\bibliography{egbib}
}

\appendix
\section{Appendix}
Table \ref{tab:imagenet-taxonomy}
describes the taxonomy of the Imagenet subset,
named ImageNet-Subtree,
which is performed for hierarchical classification in Section \ref{exp:taxonomy}.
We also provide performance curves and implementation details
as well as activation maps for NestedNet in the following sections.

\subsection{Performance Curves}\label{sec:supp-2}

\begin{figure}[t]
    \centering
    \subfigure[Train]
    {\includegraphics[width=0.48\textwidth]{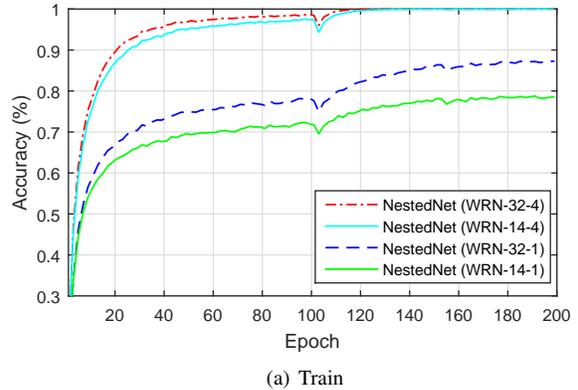}
    \label{fig:kd_curve_train}}
    \subfigure[Test]
    {\includegraphics[width=0.48\textwidth]{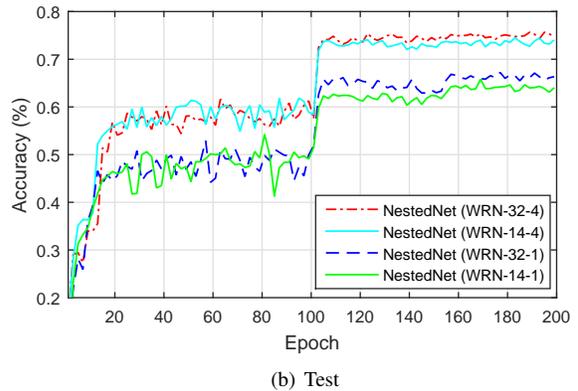}
    \label{fig:kd_curve_test}}
    \caption{
    Performance of NestedNet while training on the CIFAR-100 dataset \cite{krizhevsky2009learning}.
    ($\cdot$) denotes the applied architecture.}
    \label{fig:kd_cifar100_training}
\end{figure}

We provide performance curves of NestedNet for train and test sets in CIFAR-100
while training on the knowledge distillation problem,
where we use the same architecture to those used in Section \ref{sec:exp}.
Train and test accuracies of each internal network while learning the nested network,
which are computed in every epoch by averaging for all batch sets in train and test images, respectively,
are shown in Figure \ref{fig:kd_cifar100_training}.
For the experiment, we have empirically found that the curves
obtained from the $n$-in-1 nested sparse network,
whose internal networks are learned simultaneously,
give similar trend to those obtained from the independently learned baseline networks.
Further details and results of the nested network are described in Section \ref{exp:kd}.

\subsection{Implementation Details}\label{sec:supp-1}

\subsubsection{CIFAR datasets}
We implement NestedNets
based on state-of-the-art networks
such as residual networks (ResNet) \cite{he2016deep}
and wide residual networks (WRN) \cite{zagoruyko2016wide}.
We follow the practice in \cite{he2016deep} to construct those networks
whose number of layers is 6$n_b$+2,
where $n_b$ is the number of residual blocks.
We initialize weights in all compared architectures
using the Xavier initialization \cite{glorot2010understanding}
and train them from scratch.
For NestedNet,
we use the SGD optimizer with momentum of 0.9 and the Nesterov acceleration method
where the size of a mini-batch is 128.
Batch normalization \cite{ioffe2015batch} is adopted after each convolutional operation
and dropout \cite{srivastava2014dropout} is not used.
The learning rate starts from 0.1 and is divided by 10 when the number of iterations reaches 40K and 60K, respectively,
and the total number of iterations is 80K.
We use a standard weight decay of 0.0002.
The nested structure is implemented in all layers for adaptive deep compression
and in all residual blocks except the first convolutional and the last fully-connected layers
for the rest of the applications
based on the aforementioned architectures,
where we learn different fully-connected weights in the final layer
to address different purposes (e.g., different output dimensionality for hierarchical classification).

\begin{table*}[t]
\caption{
Taxonomy of the ImageNet-Subtree dataset.
($\cdot$) denotes the number of subclasses for each intermediate category.}

\centering
    \renewcommand\arraystretch{1.1}
    \small
    \begin{tabular}{ | c || c || c | } \hline
     Superclass  & Intermediate class  &  Subclass  \\ \hline \hline
     Natural object & Fruit (9)&  Strawberry, Orange, Lemon, Fig, Pineapple, Banana, Jackfruit, Custard apple, Pomegranate \\ \hline
     \multirow{3}{*}{Plant}    & \multirow{2}{*}{Vegetable (9)} & Head cabbage, Broccoli, Cauliflower, Zucchini, Spaghetti squash, \\
                               &  & Acorn squash, Butternut squash, Cucumber, Artichoke \\ \cline{2-3}
                               & Flower (3)    & Daisy, Yellow lady's slipper, Cardoon \\ \hline
     \multirow{8}{*}{Animal}   & \multirow{3}{*}{Dog (14)}  & Siberian husky, Australian terrier, English springer, Walker hound, Weimaraner,  \\
                               &                       & Soft coated wheaten terrier, Old English sheepdog, French bulldog, Basenji,   \\
                               &                       & Bernese mountain dog, Maltese dog, Doberman, Boston bull, Greater Swiss mountain dog \\ \cline{2-3}
                               & Cat (5)  & Egyptian cat, Persian cat, Tiger cat, Siamese cat, Madagascar cat \\  \cline{2-3}
                               & \multirow{2}{*}{Fish (10)} & Great white shark, Tiger shark, Hammerhead, Electric ray, Stingray, \\
                               &                       & Barracouta, Coho, Tench, Goldfish, Eel \\  \cline{2-3}
                               & \multirow{2}{*}{Bird (10)} & Goldfinch, Robin, Bulbul, Jay, Bald eagle, Vulture, Peacock, \\
                               &                       & Macaw, Hummingbird, Black swan \\ \hline
     \multirow{7}{*}{Artifact} & Instrument (10)  & Grand piano, Drum, Maraca, Cello, Violin, Harp, Acoustic guitar, Trombone, Harmonica, Sax \\ \cline{2-3}
                               & \multirow{2}{*}{Vehicle (10)} & Airship, Speedboat, Yawl, Trimaran, Submarine, Mountain bike, Freight car,  \\
                               &                          & Passenger car, Minivan, Sports car \\ \cline{2-3}
                               & \multirow{2}{*}{Furniture (10)} & Park bench, Barber chair, Throne, Folding chair, Rocking chair, Studio couch, Toilet seat, \\
                               &                            & Desk, Pool table, Dining table \\ \cline{2-3}
                               & \multirow{2}{*}{Construction (10)} & Suspension bridge, Viaduct, Barn, Greenhouse, Palace, Monastery, Library,  \\
                               &                               & Boathouse, Church, Mosque \\ \hline
    \end{tabular}
    \label{tab:imagenet-taxonomy}
\end{table*}

\subsubsection{ImageNet dataset}

NestedNet was constructed based on the ResNet-18 architecture
following the instruction in \cite{he2016deep} for the ImageNet dataset,
where the numbers of channels in the core and medium level networks
were set to quarter and half of the number of all channels, respectively,
for every convolutional layer.
We set different fully-connected layers in the last output layer
as performed in the CIFAR datasets.
We use the SGD optimizer with momentum of 0.9 and the Nesterov method
with the size of a mini-batch of 256 and the weight decay of 0.0001.
The learning rate starts from 0.1 and divided by 10
when the number of epochs reaches 15K, 30K, and 45K, respectively,
and the total number of epochs is 50K.
For the dataset, we learn nested parameters
sequentially from core to full level for every iteration
instead of learning them simultaneously.

\subsubsection{Consensus in NestedNet}
For NestedNet-L
described in Section \ref{exp:consensus},
which incorporates multiple knowledge from all nested levels,
we add a fully-connected layer,
called a consensus layer,
to the concatenated vector of all outputs in NestedNet,
and the consensus layer again produces an output vector whose size is the number of classes.
Note that the consensus layer is learned after NestedNet is trained in this work,
but we can learn the whole network including the consensus layer simultaneously.
When we address the hierarchical classification problem for the CIFAR-100 dataset \cite{krizhevsky2009learning}
in Section \ref{exp:taxonomy},
rather than just concatenating the two level outputs,
we collect additional fine class output (whose dimensionality is 100) in the core level network,
which requires another fully-connected layer in the final layer in NestedNet
to produce an output of different dimensionality,
and then learn the consensus layer using concatenation of the three outputs
(two fine class outputs from both full and core levels and
one coarse class output from the core level) for better prediction.
We also average two fine class outputs from both level networks to build NestedNet-A
for the hierarchical classification problem.
For more accurate inference, one can append more layers
with nonlinearity in the top of the network,
while this practice only adds a layer without nonlinearity
which may not achieve further performance gain for a certain problem.
For ImageNet, we constructed two consensus variants of NestedNet
in a similar way for CIFAR-100.
When handling the knowledge distillation problem,
we use the designed number of output features learned from NestedNet
to construct NestedNet-A and NestedNet-L.
We use the SGD optimizer without momentum for
both knowledge distillation and hierarchical classification in learning NestedNet-L.
To yield the best performance,
the learning rate for all consensus layers
starts from 0.1 and is divided by 10 when the number of iterations reaches 20K, 30K and 40K, respectively,
and the total number of iterations is set to 50K.

\begin{figure*}[t]
    \centering
    {\includegraphics[width=0.96\textwidth]{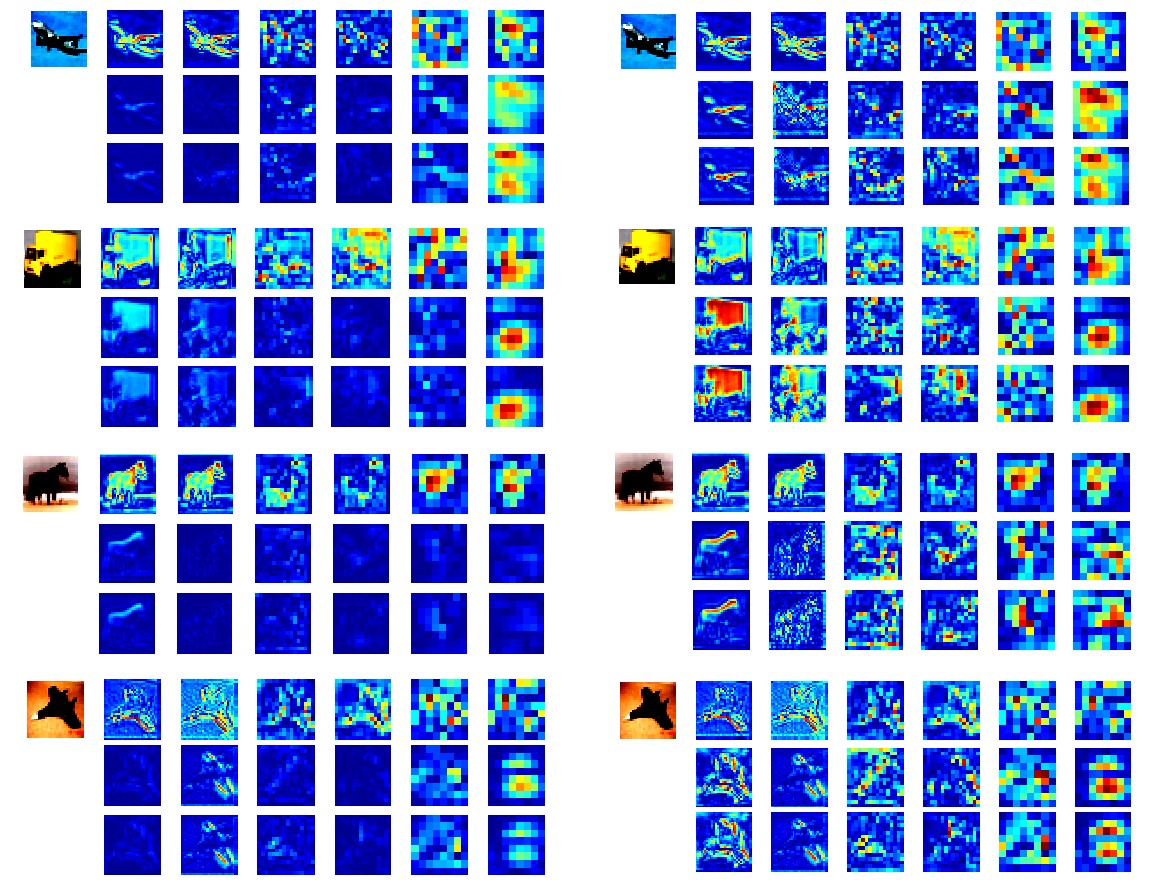}}
    \caption{
    Activation maps in NestedNet according to different layers
    on train (first) and test (second to fourth) images
    for the CIFAR-10 dataset \cite{krizhevsky2009learning}.
    The maps scaled consistently in layer-wise (left)
    and the individual maps without keeping the consistent scale (right) are provided for better understanding.
    Each row for every image represents the images in each internal network,
    from core-level (top) to full-level (bottom).
    Best viewed in color.}
    \label{fig:nsn_act_supp}
\end{figure*}

\subsection{Activation Feature Maps}\label{sec:supp-3}

Figure \ref{fig:nsn_act_supp} shows activation feature maps,
which are outputs from different layers in NestedNet
when feeding train and test images of the CIFAR-10 dataset \cite{krizhevsky2009learning} to the learned network.
Each row represents the maps obtained in each network with different nested level
from core-level (top) to full-level (bottom).
Note that the maps illustrated here are printed out using the filters in the current level network,
which do not include the filters already computed in the sub-level networks,
to see what filters in higher level networks learn
(i.e., increments from the sub-level networks).
We also provide additional activation maps for the same images
that show the individual images without keeping the consistent scale
when drawing the figures (right column in the figure),
since the filters in higher level networks sometimes produce small values
which are difficult to observe as shown in the left column of the figure.
From the figure,
we can see that the learned filters in the higher level networks
also catch the important and complementary features,
even though they are marginal compared to those in the core level network.

\end{document}